%% file: main.tex
\documentclass[10pt,twocolumn,letterpaper]{article}

\usepackage[pagenumbers]{cvpr} 
\usepackage{multirow}
\usepackage{booktabs}
\usepackage{adjustbox}
\usepackage{amsmath,amssymb}
\usepackage{microtype}
\usepackage{graphicx}

\usepackage{caption}

\definecolor{cvprblue}{rgb}{0.21,0.49,0.74}
\usepackage[pagebackref,breaklinks,colorlinks,allcolors=cvprblue]{hyperref}
\usepackage{color}

\def\paperID{} 
\def\confName{CVPR}
\def\confYear{2026}

\title{Pareto-Guided Optimization for Uncertainty-Aware Medical Image Segmentation} 

\author{Jinming Zhang$^{1}$\quad
Youpeng Yang$^{2}$\quad
Xi Yang$^{1}$\quad
Haosen Shi$^{1}$\\
Yuyao Yan$^{1}$\quad
Qiufeng Wang$^{1}$\quad
Guangliang Cheng$^{3}$\quad
Kaizhu Huang$^{4}$\\[0.5em]
$^{1}$Xi'an Jiaotong-Liverpool University\quad
$^{2}$Zhejiang University\\
$^{3}$University of Liverpool\quad
$^{4}$Duke Kunshan University
}

\begin{document}

\twocolumn[
\maketitle
\begin{center}
\resizebox{\linewidth}{!}{
\includegraphics{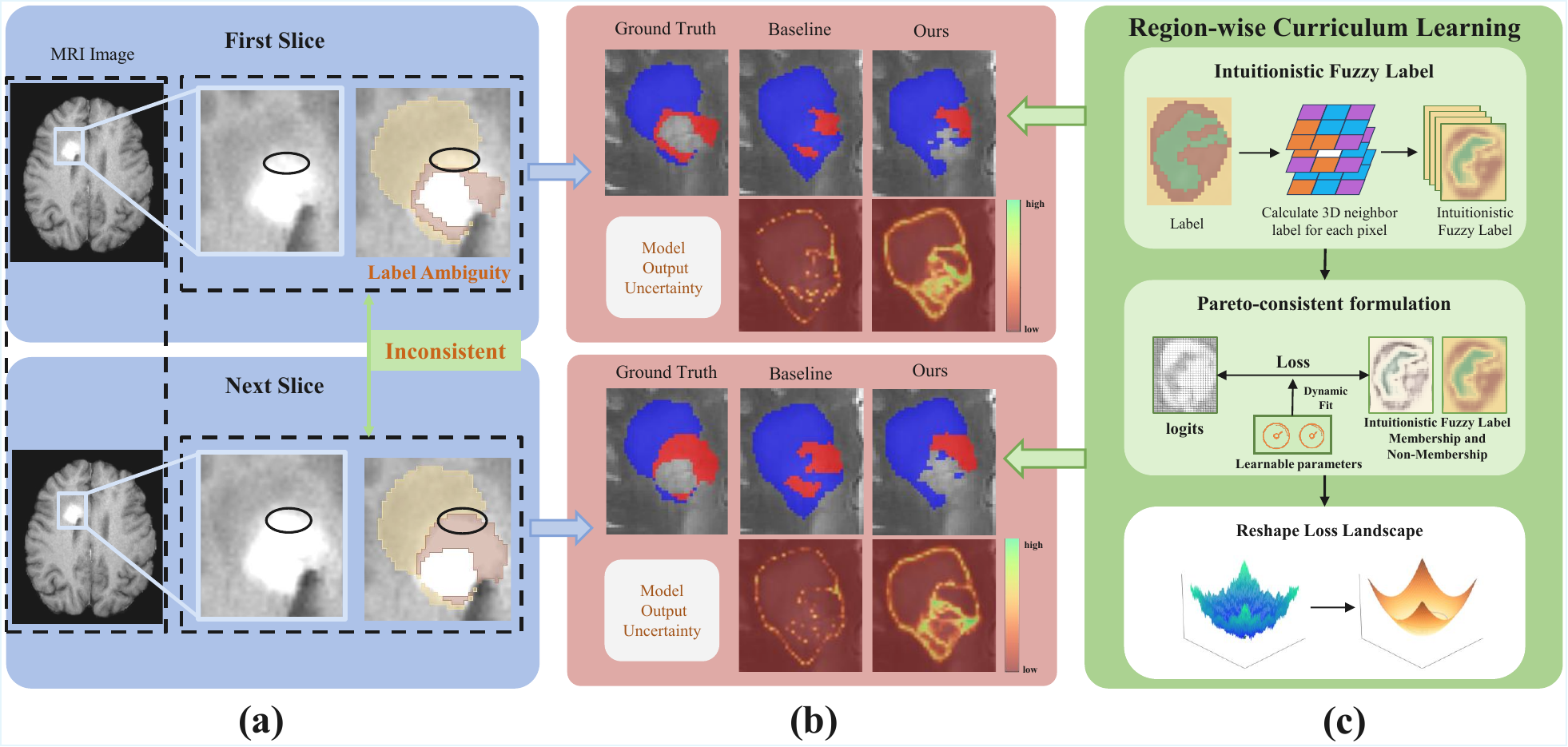}
}
\captionof{figure}{
The proposed region-aware segmentation framework that tackles boundary ambiguity and slice-wise inconsistencies through (a) label ambiguity and label inconsistency between adjacent slices, (b) model output and uncertainty map, and (c) a region-wise curriculum learning approach with intuitionistic fuzzy labels to stabilise training and improve segmentation accuracy.
}
\label{fig:banner}
\end{center}
]


\input{sec/0_abstract}    
\input{sec/1_intro}
\input{sec/2_relatedwork}

\input{sec/3_methodology}

\input{sec/4_experiments}
{
    \small
    \bibliographystyle{ieeenat_fullname}
    \bibliography{main}
}

\input{sec/X_suppl}

\end{document}

%% file: sec/0_abstract.tex
\begin{abstract}
Uncertainty in medical image segmentation is inherently non-uniform, with boundary regions exhibiting substantially higher ambiguity than interior areas.
Conventional training treats all pixels equally, leading to unstable optimization during early epochs when predictions are unreliable.
We argue that this instability hinders convergence toward Pareto-optimal solutions and propose a region-wise curriculum strategy that prioritizes learning from certain regions and gradually incorporates uncertain ones, reducing gradient variance.
Methodologically, we introduce a Pareto-consistent formulation that balances trade-offs between regional uncertainties by adaptively reshaping the loss landscape and constraining convergence dynamics between interior and boundary regions; this guides the model toward Pareto-approximate solutions.
To address boundary ambiguity, we further develop a fuzzy labeling mechanism that maintains binary confidence in non-boundary areas while enabling smooth transitions near boundaries, stabilizing gradients, and expanding flat regions in the loss surface.
Extensive evaluations on the BraTS18 and Pretreat-MetsToBrain-Masks tumor segmentation benchmarks demonstrate that our approach consistently enhances segmentation performance and training stability compared to traditional crisp-set methods. Our method excels in both conventional and challenging scenarios with missing or single modalities, demonstrating robustness and adaptability across tasks.

\end{abstract}

%% file: sec/1_intro.tex
\section{Introduction}
\label{sec:intro}

Medical image segmentation is crucial for diagnosis and treatment but faces uncertainty and instability due to low contrast, weak gradients, and inconsistent annotations~\cite{li2024boundary,mehrtash2020confidence,jungo2020analyzing,litjens2017survey,hesamian2019deep}. These issues can mislead optimization, hinder reliable decision-making, and cause inaccurate predictions. Previous studies have delineated uncertainty into two distinct components: aleatoric and epistemic.  \cite{zou2023review,jones2022direct,ahdritz2024provable,kendall2017uncertainties,abdar2021review,hullermeier2021aleatoric,der2009aleatory}. 
In medical imaging, aleatoric uncertainty primarily arises from inherent intrinsic ambiguity and acquisition-related noise, making it largely irreducible by additional data alone. Epistemic uncertainty reflects the model’s limited knowledge and is pronounced under data scarcity, class imbalance, and domain shifts across institutions. In medical image segmentation, experts' annotation inconsistencies are typically treated as aleatoric because they reflect genuine ambiguity in the ground truth; however, when inconsistency primarily stems from non-standardized labeling protocols or insufficient expertise, it also carries an epistemic component, which tends to diminish once consensus guidelines are adopted \cite{hu2019supervised,jungo2019assessing,reinke2024understanding}. 

As illustrated in Fig.~\ref{fig:banner} (a), label ambiguity arises from inconsistencies between adjacent slices, where one slice is mislabeled (highlighted in the top row). 
However, current uncertainty modeling strategies in medical imaging primarily focus on pixel-level aleatoric or heteroscedastic uncertainties~\cite{wang2019aleatoric,kuestner2024predictive,zhang2024heteroscedastic}, often overlooking the uncertainty originating within the label space itself. Segmentation labels inherently carry spatially varying uncertainty, as regions like boundaries and interiors differ in reliability due to imprecise delineation and supervision. We argue that \textbf{such intra-sample label-driven heterogeneity is a fundamental yet underexplored uncertainty source, crucial for enhancing segmentation stability and reliability.}

Based on empirical observations, interior regions typically exhibit low ambiguity and stable gradients, while boundary regions show greater uncertainty due to overlap between foreground and background representations. Pareto analysis indicates that focusing on these high-uncertainty regions during initial training phases can destabilize the optimization process, leading to segmentation errors or missing structures even in low-uncertainty areas. This highlights how conventional training that treats all pixels equally can exacerbate label ambiguity early in learning.   

These findings motivate a region-adaptive training approach for segmentation models, termed a \textit{Region-wise Curriculum Learning Strategy}, emerging naturally from Pareto optimization principles.
Instead of treating all label regions uniformly, the model initially focuses on stable, low-uncertainty interior regions and gradually extends attention to high-uncertainty boundaries. This progressive approach stabilizes gradient flow, reducing variance and preventing early-stage instability. As shown in Fig.~\ref{fig:banner} (b), an uncertainty map, constructed by the maximum value in the model’s softmax logits for each pixel, effectively identifies regions with high data uncertainty. This enables targeted learning in regions with label ambiguity (as shown in Fig.~\ref{fig:banner} (a)), resulting in improved segmentation performance and verified stabilization of the training dynamics.

To implement the curriculum mechanism, we introduce an Intuitionistic Fuzzy Label (IFL) representation that models uncertain boundaries through soft labeling while preserving hard assignments in confident regions to ensure stable supervision. Specifically, IFL assigns each pixel a membership degree based on its local neighborhood, enabling smooth transitions across uncertain boundaries and maintaining high confidence in interior regions. When integrated into Dice and an auxiliary loss formulations, this approach flattens the loss surface around ambiguous regions, allowing for smoother optimization and greater robustness \cite{muller2019does}.
Furthermore, uncertainty heterogeneity across regions introduces conflicting learning dynamics between stable interior areas and ambiguous boundaries. To address these conflicts, we propose a Pareto-consistent formulation embedded directly into an Fuzzy Auxiliary Loss. Instead of relying solely on external weighting or projection mechanisms, this formulation complements them with a continuous internal adjustment process. By using two learnable parameters ($\rho_1$, $\rho_2$), the loss landscape dynamically adjusts, enabling fine-tuned convergence behaviors between full and boundary regions~\cite{kumar2010self}. This facilitates the model in approximating Pareto-optimal solutions (see Fig.~\ref{fig:banner}(c)).

\noindent Our main contributions are as follows:
\begin{itemize}
    \item We propose a \textbf{Region-wise Curriculum Learning Strategy} with a \textbf{Pareto-consistent formulation} to effectively approximate Pareto-optimal solutions, enhancing segmentation robustness and stabilizing training.
    \item We pioneer the explicit modeling of label uncertainty througha  pixel-wise \textbf{IFL representation}, enabling the model to differentiate between boundary and non-boundary regions effectively. 
    \item Extensive evaluations on two medical segmentation benchmarks show our approach enhances segmentation performance and training stability over crisp-set baselines, especially in conventional segmentation, missing-modality, and single-modality tasks.
\end{itemize}

%% file: sec/2_relatedwork.tex
\section{Related Work}
\label{sec:relatedwork}

\paragraph{Uncertainty in Segmentation} Uncertainty in segmentation involves aleatoric (data noise) and epistemic (model limitations) types \cite{kendall2017uncertainties, gal2016dropout, ahdritz2024provable}. Aleatoric is irreducible; epistemic can be reduced with diverse data or better models \cite{kumar2010self}. Noisy labels can misclassify clean samples, leading to overconfidence and less reliable uncertainty estimates \cite{xia2021sample,karimi2020deep,sukhbaatar2014learning,northcutt2021confident}. In multi-view classification, they cause overfitting and reduce uncertainty for clean samples~\cite{xu2025noisy, han2018co,li2020dividemix,xu2024trusted}. Rank loss and confidence-based sample selection address overconfidence \cite{wang2021rethinking,muller2019does}. Uncertainty varies within a sample: boundary regions have higher aleatoric uncertainty due to noise; interior areas are more certain \cite{wang2019aleatoric,kuestner2024predictive,zhang2024heteroscedastic,baumgartner2019phiseg,jones2022direct}. Current methods often overlook label uncertainty, essential in segmentation tasks.

\paragraph{Curriculum Learning} Introduced by Bengio et al.~\cite{bengio2009curriculum}, Curriculum Learning (CL) uses an easy to hard training strategy to enhance model convergence and generalization. The optimal difficulty measure is inherently task-specific. While applied in medical segmentation by ranking samples based on difficulty~\cite{peng2021self,zhou2017fixed}, existing methods operate at the sample level, overlooking region-wise differences in label reliability.

\paragraph{Pareto Optimization} Adaptive weighting and gradient-based balancing methods are well-studied. Kendall et al.~\cite{kendall2018multi} proposed weighting tasks by their predictive uncertainty for automatic loss balancing. Sener and Koltun~\cite{sener2018multi} viewed multi-task learning as multi-objective optimization with MGDA for Pareto front approximation. GradNorm adjusts gradients to equalize task learning rates~\cite{chen2018gradnorm}. Recent advances extend these concepts to explicitly approximate Pareto sets in deep networks~\cite{lin2022pareto,ruchte2021scalable,tuan2024framework}.
While effective, current Pareto optimization techniques mainly focus on loss-level adjustments. This external control can be coarse, causing instability or oscillations when objectives conflict. Such adjustments may lead to suboptimal balances, especially in scenarios with multiple, interacting objectives.

\paragraph{Fuzzy Set Theory}
Fuzzy set theory naturally models uncertainty by assigning a continuous membership degree to each element, ranging from 0 to 1~\cite{zadeh1965fuzzy}. Intuitionistic fuzzy sets further capture uncertainty by incorporating a hesitation degree, which is useful for distinguishing data near class boundaries~\cite{atanassov1986intuitionistic,yang2024negative}.Unlike hard labels that assume distinct class boundaries, intuitionistic fuzzy sets offer a soft representation, ideal for segmentation tasks with ambiguous boundaries or noisy annotations.  Traditional approaches often apply fuzzy modeling to the data, treating fuzziness as a feature space property~\cite{clairet2006color,pagola2012interval}. However, this perspective overlooks variability in uncertainty across the label space. Our approach constructs pixel-wise fuzzy membership functions on segmentation labels, enabling networks to learn spatially adaptive uncertainty representations that more accurately capture annotation ambiguity.

%% file: sec/3_methodology.tex
\section{Methodology}
\label{sec:method}

\subsection{ Intuitionistic Fuzzy Label} \label{sec:uncertainty}

\paragraph{Aleatoric and Epistemic Uncertainty} 
Given input image $\mathbf{x} \in \mathcal{X}$ and label map $\mathbf{y} \in \mathcal{Y}$, where each pixel $y_i \in \{1,2, \dots, C\}$ corresponds to one of $C$ classes, the predictive distribution is $p_\theta(\mathbf{y} | \mathbf{x})$ parameterized by $\theta$. The predictive uncertainty for pixel $i$ is decomposed as follows\footnote{Note: Some works adopt the opposite naming convention; here the terms follow~\cite{kendall2017uncertainties}.}:
\begin{equation}
\label{eq:uncertainty-decomposition}
\begin{split}
    \mathrm{Var}_{p(\mathbf{y}|\mathbf{x})}[y_i]
    & = \underbrace{\mathbb{E}_{p(\theta|\mathcal{D})}[\mathrm{Var}_{p(\mathbf{y}|\mathbf{x},\theta)}[y_i]]}_{\text{Aleatoric Uncertainty}} \\ 
    & \quad + \underbrace{\mathrm{Var}_{p(\theta|\mathcal{D})}[\mathbb{E}_{p(\mathbf{y}|\mathbf{x},\theta)}[y_i]]}_{\text{Epistemic Uncertainty}},
\end{split}
\end{equation}
where $\mathcal{D}$ denotes the training data distribution. 

In medical image segmentation, boundary regions between classes often exhibit high aleatoric uncertainty from partial volume effects or imprecise annotations. Epistemic uncertainty indicates the model's confidence in its predictions and may rise when the model overfits these inherently ambiguous boundaries. 
Soft labels represent uncertainty levels within the labels, capturing the boundary ambiguity, aiding in modelling aleatoric uncertainty. By reparameterizing the predictive distribution $p_\theta(\mathbf{y}|\mathbf{x})$, they convert pixel labels $y_i$ into probability distributions, rather than relying on hard labels. This reduces overfitting and decreases epistemic uncertainty.


\paragraph{Construction of Fuzzy Labels}
This study employs soft labels via intuitionistic fuzzy sets, establishing probability distributions that encompass both membership and non-membership degrees for each pixel across different classes, offering a nuanced representation of uncertainty.  Additionally, the incorporation of hesitation degrees for each class enhances the ability to effectively separate boundary regions. 
A fuzzy label $\tilde{y}$ is constructed for each pixel (or voxel), representing a continuous membership degree rather than a crisp one hot label.

Given a binary ground truth label $y \in \{0, 1\}$, the fuzzy membership $\mu(x)$ and non-membership $\nu(x)$ for pixel $x$ represent the belongingness or not to the specific class, respectively. They satisfy the following constraint:
\begin{equation}
0\leq \mu(x) + \nu(x) \leq 1.
\end{equation}

The hesitation degree $\pi(x)$ notes the complementary part of membership and non-membership degrees in the total information, which implies:
\begin{equation}\label{eqn_hesi}
    \pi(x) = 1- \mu(x) - \nu(x).
\end{equation}
The membership degree $\mu(x)$ is estimated based on the spatial consistency of neighboring pixels belonging to the same class. Let $\mathcal{N}(x)$ denote a local neighborhood centered at $x$. The fuzzy membership is computed as:
\begin{equation}
\mu(x) = \frac{1}{|\mathcal{N}(x)|} \sum_{x' \in \mathcal{N}(x)} \mathbb{I}(y_{x'} = y_x),
\end{equation}
where $\mathbb{I}(\cdot)$ is the indicator function. Intuitively, pixels surrounded by homogeneous labels obtain $\mu(x) \approx 1$, while pixels located at class boundaries yield intermediate values, thus encoding the degree of boundary uncertainty.

The non-membership degree $\nu(x)$ represents the degree to which pixel $x$ is not belong to a given class. It is calculated as the complement of the membership degree $\mu(x)$, adjusted by a learnable parameter $\rho_2$ for adaptive uncertainty control:
\begin{equation}
\nu(x) = \left( 1 - \mu(x) \right) \cdot \rho_2.
\end{equation}
where $\rho_2$ is a learnable scaling factor constrained within [0, 1], used to adjust the influence of the non-membership degree during training and modulate uncertainty.



\subsection{Pareto-consistent Formulation}
\paragraph{Fuzzy Auxiliary loss and Boundary Smoothness}
Given the fuzzy membership label $\mu_c(x)$ and its complementary non-membership label $\nu_c(x)$ defined in Sec.~\ref{sec:uncertainty}, 
let $p_c(x) = \mathrm{softmax}(z_c(x))$ denote the predicted probability for class $c$.
The proposed fuzzy auxiliary loss is formulated as:
\begin{equation}
\begin{split}
\mathcal{L}_{\text{fuzzy}} = & 
- \sum_{c=1}^{C} 
\big( 
\mu_c(x) \log p_c(x)\\
&  + \rho_2 (1-\mu_c(x)) \log \rho_1 (1 - p_c(x))
\big),
\label{eq:fuzzy_loss}
\end{split}
\end{equation}
where $\rho_1, \rho_2 \in (0,1)$ are two learnable scalars obtained via sigmoid reparameterization
$\rho_i = \sigma(\tilde{\rho}_i)$. 
$\rho_2$ modulates the credibility of the fuzzy label itself, while $\rho_1$ controls the penalty strength imposed on the predicted logits.
To further reveal the smoothing property of the fuzzy loss near object boundaries, 
we analyze its gradient with respect to $p_c(x)$:
\begin{equation}
\frac{\partial \mathcal{L}_{\text{fuzzy}}}{\partial p_c}
= -\frac{\mu_c(x)}{p_c(x)} 
+ \frac{\rho_2 (1 - \mu_c(x))}{1 - p_c(x)}.
\label{eq:grad_p}
\end{equation}
Therefore, $\rho_1$ governs the sharpness of the model response, while $\rho_2$ regulates the trust placed on soft boundary labels. In contrast to standard cross entropy or Dice loss, which impose sharp transitions when $\mu_c \in \{0,1\}$,
our formulation introduces an adaptive slope in the range $\mu_c \in (0,1)$, 
corresponding to pixels near class boundaries.
This adaptivity effectively smooths the loss landscape and reduces oscillatory gradients in ambiguous regions, 
thereby improving convergence stability and segmentation consistency.

To interpret their effect, we analyze the gradients of Eq.~(\ref{eq:fuzzy_loss}) with respect to $\rho_1$ and $\rho_2$:
\begin{equation}
\begin{split}
\frac{\partial \mathcal{L}_{\text{fuzzy}}}{\partial \rho_1}
&= - \sum_{c} \frac{\rho2}{\rho1} (1-\mu_c(x)), \\
\frac{\partial \mathcal{L}_{\text{fuzzy}}}{\partial \rho_2}
&= - \sum_{c} (1-\mu_c(x)) \log \rho_1(1-p_c(x)).
\end{split}
\label{eq:grad_rho}
\end{equation}
$\rho_1$ primarily adjusts the sharpness of the model response by scaling the penalty on confidently predicted regions, 
while $\rho_2$ regulates the penalty strength of uncertain (low membership) or boundary regions.


In non-boundary regions where $\mu_c \in \{0,1\}$, the fuzzy loss degenerates to the main loss, introducing no additional modulation. 
In contrast, for boundary regions where $\mu_c \in (0,1)$, the fuzzy loss introduces a hierarchical weighting mechanism. 
Specifically, $\rho_2$ controls the credibility of soft labels, thereby determining the relative confidence between soft and hard boundaries; 
$\rho_1$, in turn, regulates the model’s response strength to these boundary pixels by shaping the gradient slope with respect to $p_c(x)$. 
Through this coupling, the fuzzy loss \textbf{first adjusts the balance between soft and hard boundaries, and subsequently alters the overall weighting of boundary versus non-boundary regions}. 
In essence, the fuzzy loss achieves a global rebalancing effect by locally modulating the trust in soft boundaries making boundary supervision smoother and more adaptive without explicitly defining multiple objectives.

\subsection{Region-wise Curriculum Learning Strategy} \label{sec:Curriculum}

\paragraph{Overall Loss and Training Dynamics}
The overall optimization objective evolves with time as:
\begin{equation}
\mathcal{L}(\theta, t) = 
\mathcal{L}_{\text{Dice}}(\theta) + \lambda(t)\, \mathcal{L}_{\text{fuzzy}}(\theta,\rho_1, \rho_2),
\label{eq:time_loss}
\end{equation}
where $\mathcal{L}_{\text{Dice}}$ denotes the Dice loss and $\mathcal{L}_{\text{fuzzy}}$ represents the fuzzy auxiliary loss.  The coefficient $\lambda(t)$ can either follow a predefined schedule (e.g., exponential or linear decay) or be implicitly realized by the learnable parameter $\rho_1$, whose adaptation produces a similar effect.


From a Pareto optimization viewpoint, $\mathcal{L}_{\text{Dice}}$ and $\mathcal{L}_{\text{fuzzy}}$
represent two competing objectives regional precision and boundary adaptability. 
Each weight $\lambda(t)$ corresponds to a specific trade-off point on the Pareto frontier.
Gradually decreasing $\lambda(t)$ 
induces a continuous trajectory along this frontier,
shifting the optimization from a fuzzy dominant regime to a main loss dominant regime.
Unlike discrete reweighting between multiple losses,
this approach achieves an implicit Pareto traversal within a single differentiable objective.

\paragraph{Continuous Pareto Dynamics}
The optimization dynamics can be expressed as:
\begin{equation}
\dot{\theta}(t) = - \nabla_{\theta} \mathcal{L}(\theta,t)
               = - \nabla \mathcal{L}_{\text{Dice}}
                 - \lambda(t)\, \nabla \mathcal{L}_{\text{fuzzy}}.
\end{equation}
As $\lambda(t)$ decays, the optimization direction gradually aligns with 
$-\nabla \mathcal{L}_{\text{Dice}}$, 
indicating a transition from exploration (boundary-aware learning)
to exploitation (region-precision refinement).
This can be viewed as a constrained trajectory along the Pareto manifold 
in the objective space $(\mathcal{L}_{\text{main}}, \mathcal{L}_{\text{fuzzy}})$,
where early training explores broader optima to escape local minima,
and later stages focus on convergence stability.

\paragraph{Practical Realization}
Two complementary strategies are implemented:
(1) a fixed decaying schedule $\lambda(t) = \lambda_0 \exp(-\alpha t)$;
and (2) a fully learnable configuration where $(\rho_1, \rho_2)$ 
are initialized at higher values and optimized jointly with network parameters.  
The former provides a controlled analytical trajectory, 
while the latter yields a data-driven Pareto path with adaptive flexibility.
In Sec.~4, we visualize the evolution of $\lambda(t)$ and $\rho_1(t)$
along with the corresponding Pareto points, 
demonstrating how the training dynamically moves from boundary exploration to precision convergence.

\paragraph{Pareto Insight}
Unlike conventional dynamic weighting methods that adjust $\lambda(t)$ externally,
our formulation embeds two learnable parameters $\rho_1$ and $\rho_2$ \emph{inside} the loss itself.
This design allows the optimization process to adaptively reshape the loss landscape in response to $\lambda(t)$.
Specifically, as $\lambda(t)$ decreases, the influence of the fuzzy loss diminishes, 
and $\rho_2$ passively adjusts the credibility of the soft boundary labels generated from the fuzzy set. 

Through this adaptive interaction, $\rho_2$ implicitly establishes a Pareto relationship between the soft boundary label which is from the fuzzy membership and the hard boundary label which is from the Dice loss, regulating the overall representation of boundary regions. 
The updated boundary representation then interacts with the non-boundary region through another Pareto relation, 
achieving a second-level equilibrium that balances the contributions of boundary and non-boundary areas. 
Overall, $\lambda(t)$ acts as an active controller that gradually shifts the optimization focus from fuzzy regularization to hard supervision, 
while $\rho_2$ serves as a responsive modulator that realizes a hierarchical Pareto balance between soft hard boundaries and between boundary non-boundary regions.

%% file: sec/4_experiments.tex
\section{Experiments}
\label{sec:experiments}

\subsection{Setup}

\paragraph{Datasets}  
The proposed method is evaluated using two 3D medical image segmentation benchmarks.
{BraTS18} provides  multi-modal MRI scans of non-metastatic brain tumors, with four modalities (T1, T1ce, T2, FLAIR).  
{Pretreat-MetsToBrain-Masks}  targets metastatic brain tumors and is pre-processed to align voxel spacing and intensity normalization with those of the BraTS18 data.  
For both datasets, we adopt a patch-based training scheme, cropping volumetric regions of size $96\times128\times128$ voxels, with stride equal to half the patch size in each dimension.  
The fuzzy label neighbourhood radius is set to $r=1$, corresponding to $26$ immediate neighbours in 3D space.

\paragraph{Models}  
The approach is validated against three representative 3D segmentation architectures:  
\begin{itemize}  
  \item {\bf mmFormer}~\cite{zhang2022mmformer}, a multimodal medical transformer for brain tumor segmentation with missing MRI modalities.
    \item {\bf SwinUNETR}~\cite{hatamizadeh2021swin}, 3D Swin-Transformer encoder with UNet style decoder for brain tumor segmentation.  
    \item {\bf VNet}~\cite{milletari2016v}, a volumetric fully convolutional network for 3D medical image segmentation.
\end{itemize} 

VNet was trained and evaluated in a single-modality setting to assess its performance under this extremely constrained training regime.
The other networks and tasks were tested with full and missing modalities conditions. For all experiments, we kept the original model losses and incorporated our method to improve performance across these diverse scenarios. 
Some modality configurations were evaluated on a subset of networks due to computational limits.

\paragraph{Implementation Details}  
Training is conducted on a heterogeneous GPU platform comprising 4 × RTX 3090 cards. We use the Adam optimizer with weight decay set to $10^{-5}$. The initial learning rate is set to $1\times10^{-3}$, and batch size is chosen depending on GPU memory (usually 4).  
For loss composition, each baseline retains its original loss function, and is further augmented with our method. In our experiments we further include the fuzzy loss module with two trainable parameters $\rho_{1}$, $\rho_{2}$. 
Training proceeds for a maximum of 1000 epochs with a cosine decay learning rate schedule. We report region-wise Dice Score: whole tumor, tumor core, enhancing tumor (wt, tc, et) and mean Dice Score. 

\paragraph{Rationale for Network–Modality Pairing}
Due to the substantial computational cost of 3D medical segmentation, 
we conducted a subset of the possible network–modality combinations. 
The main paper reports representative results from three complementary settings, 
Additional experiments and ablations are provided in the supplementary material, due to space and computational constraints.
This selection provides a balanced coverage across CNN and Transformer architectures 
as well as different levels of data uncertainty.

\begin{table}
\caption{Segmentation performance (Dice scores) of mmFormer~\cite{zhang2022mmformer} and SwinUNETR~\cite{hatamizadeh2021swin} with and without the proposed method on BraTS18 (“BraTS”) and Pretreat-MetsToBrain-Masks (“Metastasis”) under the full-modality setting.}
  \label{tab:full_modalities}
  \centering
  \resizebox{\linewidth}{!}{
  \begin{tabular}{c|lcccc}
    \toprule
    Dataset & Methods & Mean & WT & TC & ET \\
    \midrule
    \multirow{4}{*}{BraTS}
    & mmFormer & 80.36 & 90.34 & 82.56 & 68.19 \\
    & mmFormer+ours & 82.24 & 90.37 & 83.26 & 73.09 \\
    \cmidrule(r){2-6}
    & SwinUNETR & 82.37 & 87.81 & 83.06 & 76.24 \\
    & SwinUNETR+ours & 83.38 & 88.92 & 83.20 & 78.01 \\
    \midrule
    \multirow{4}{*}{Metastasis}
    & mmFormer & 66.43 & 69.24 & 67.61 & 62.45 \\
    & mmFormer+ours & 66.54 & 69.29 & 67.57 & 62.76 \\
    \cmidrule(r){2-6}
    & SwinUNETR & 61.61 & 63.22 & 61.30 & 60.29 \\
    & SwinUNETR+ours & 64.79 & 67.01 & 62.54 & 64.84 \\
    \bottomrule
  \end{tabular}
  }
\end{table}

\begin{table*}[t]
\caption{
Dice scores of baseline models and their fuzzy-regularized variants on BraTS18 under missing-modality and full-modality settings.
Here, wt, tc, and et denote Whole Tumor, Tumor Core, and Enhancing Tumor, respectively.
The modality labels T1, T2, T1c, and F correspond to T1, T2, T1ce, and FLAIR, respectively, and Full indicates the full-modality setting.
}

\label{tab:swinunetr}
\centering

\resizebox{\textwidth}{!}{
\begin{tabular}{c|l|cccc|cccccc|cccc|c|c}
\toprule

 &  & \multicolumn{4}{c|}{Remain 1 modality} & \multicolumn{6}{c|}{Missing 2 modalities} & \multicolumn{4}{c|}{Missing 1 modality} & Full & Avg. \\
\midrule

 Type & Method & T1 & T2 & T1ce & Flair & T1, T2 & T1, T1c & T1, F & T2, T1c & T2, F & T1c, F & T1 & T2 & T1ce & Flair &  &  \\
\midrule
\multirow{8}{*}{WT} 
& U-HeMIS & 57.62 & 80.96 & 61.53 & 52.48 & 68.99 & 82.95 & 82.48 & 64.62 & 68.47 & 82.41 & 83.85 & 72.31 & 83.43 & 83.94 & 84.74 & 74.05\\
& U-HVED & 49.51 & 79.83 & 53.62 & 84.39 & 85.93 & 87.58 & 81.32 & 85.71 & 64.22 & 81.56 & 88.09 & 86.72 & 88.07 & 82.32 & 88.46 & 79.16\\
\cmidrule(r){2-18}
& mmFormer & 67.52 & 81.15 & 72.22 & 86.10 & 87.30 & 87.59 & 82.99 & 87.06 & 74.42 & 82.20 & 88.14 & 87.33 & 87.75 & 82.71 & 89.64 & 82.94\\
& w/ ours & \textbf{75.25} & \textbf{85.90} & \textbf{77.64} & \textbf{86.96} & \textbf{89.43} & \textbf{89.27} & \textbf{86.94} & \textbf{88.81} & \textbf{79.87} & \textbf{86.49} & \textbf{90.00} & \textbf{89.62} & \textbf{89.55} & \textbf{87.28} & \textbf{90.04} & \textbf{86.20}\\
& $\Delta$ & \textit{+7.73} & \textit{+4.75} & \textit{+5.42} & \textit{+0.86} & \textit{+2.13} & \textit{+1.68} & \textit{+3.95} & \textit{+1.75} & \textit{+5.45} & \textit{+4.29} & \textit{+1.86} & \textit{+2.29} & \textit{+1.80} & \textit{+4.57} & \textit{+0.40} & \textit{+3.26}\\
\cmidrule(r){2-18}
 & SwinUNRTR & 68.32 & 80.58 & 70.04 & 83.69 & 86.85 & 86.20 & 83.22 & 85.77 & 73.69 & 83.07 & 87.81 & 87.22 & 86.78 & 84.35 & 87.81 & 82.36\\
 & w/ ours & 71.81 & 83.29 & 72.83 & 84.15 & 87.92 & 87.35 & 84.76 & 86.90 & 75.79 & 85.14 & 88.81 & 88.83 & 88.40 & 85.70 & 88.92 & 84.04\\
 & $\Delta$ & \textit{+3.49} & \textit{+2.71} & \textit{+2.79} & \textit{+0.46} & \textit{+1.07} & \textit{+1.15} & \textit{+1.54} & \textit{+1.13} & \textit{+2.10} & \textit{+2.07} & \textit{+1.00} & \textit{+1.61} & \textit{+1.62} & \textit{+1.35} & \textit{+1.11} & \textit{+1.68}\\
\midrule
\multirow{8}{*}{TC}
& U-HeMIS & 37.39 & 57.20 & 65.29 & 26.06 & 71.49 & 57.68 & 76.64 & 41.12 & 72.46 & 60.92 & 77.53 & 76.01 & 60.32 & 78.96 & 79.48 & 62.57\\
& U-HVED & 33.90 & 54.67 & 59.59 & 57.90 & 75.07 & 62.70 & 73.92 & 61.14 & 67.55 & 56.26 & 76.75 & 77.05 & 63.14 & 75.28 & 77.71 & 64.84\\
\cmidrule(r){2-18}
& mmFormer & 56.55 & 64.20 & 75.41 & 61.21 & 77.88 & 69.75 & 78.61 & 65.91 & 78.59 & 69.42 & 79.55 & 79.80 & 71.52 & 80.39 & \textbf{85.78} & 72.97\\
& w/ ours & 61.47 & 68.06 & \textbf{81.54} & 64.51 & \textbf{83.52} & 70.87 & \textbf{84.14} & 68.81 & \textbf{82.03} & 68.96 & \textbf{83.62} & \textbf{83.80} & 70.51 & \textbf{84.35} & 83.89 & 76.01\\
& $\Delta$ & \textit{+4.92} & \textit{+3.86} & \textit{+6.13} & \textit{+3.30} & \textit{+5.64} & \textit{+1.12} & \textit{+5.53} & \textit{+2.90} & \textit{+3.44} & \textit{-0.46} & \textit{+4.07} & \textit{+4.00} & \textit{-1.01} & \textit{+3.96} & \textit{-1.89} & \textit{+3.03}\\
\cmidrule(r){2-18}
 & SwinUNRTR & 60.50 & 67.14 & 77.48 & 68.59 & 81.04 & 71.36 & 80.73 & 71.56 & 77.56 & 70.68 & 82.21 & 82.57 & 72.85 & 81.24 & 83.06 & 75.24\\
 & w/ ours & \textbf{66.02} & \textbf{70.10} & 80.09 & \textbf{69.12} & 83.10 & \textbf{71.91} & 83.06 & \textbf{72.92} & 81.70 & \textbf{72.74} & 82.90 & 83.40 & \textbf{73.63} & 83.48 & 83.20 & \textbf{77.16}\\
 & $\Delta$ & \textit{+5.52} & \textit{+2.96} & \textit{+2.61} & \textit{+0.53} & \textit{+2.06} & \textit{+0.55} & \textit{+2.33} & \textit{+1.36} & \textit{+4.14} & \textit{+2.06} & \textit{+0.69} & \textit{+0.83} & \textit{+0.78} & \textit{+2.24} & \textit{+0.14} & \textit{+1.92}\\
\midrule
\multirow{8}{*}{ET}
& U-HeMIS & 10.16 & 25.63 & 62.02 & 11.78 & 66.10 & 30.22 & 67.83 & 10.71 & 66.22 & 32.39 & 68.72 & 68.54 & 31.07 & 69.92 & 70.24 & 46.10\\
& U-HVED & 8.60 & 22.82 & 57.64 & 23.80 & 68.36 & 32.31 & 67.83 & 27.96 & 61.11 & 24.29 & 68.93 & 68.60 & 32.34 & 67.75 & 69.03 & 46.76\\
\cmidrule(r){2-18}
& mmFormer & 32.53 & 43.05 & 72.60 & 39.33 & 75.07 & 47.52 & 74.51 & 42.96 & 74.04 & 44.99 & 75.67 & 75.47 & 47.70 & 74.75 & 77.61 & 59.85\\
& w/ ours & 34.95 & 37.16 & 67.98 & 36.94 & 69.88 & 43.19 & 69.19 & 42.68 & 69.24 & 41.23 & 71.35 & 70.78 & 46.19 & 70.49 & 70.89 & 56.14\\
& $\Delta$ & \textit{+2.42} & \textit{-5.89} & \textit{-4.62} & \textit{-2.39} & \textit{-5.19} & \textit{-4.33} & \textit{-5.32} & \textit{-0.28} & \textit{-4.80} & \textit{-3.76} & \textit{-4.32} & \textit{-4.69} & \textit{-1.51} & \textit{-4.26} & \textit{-6.72} & \textit{-3.71}\\
\cmidrule(r){2-18}
 & SwinUNRTR & 37.44 & 42.53 & 72.20 & 42.37 & 75.72 & 45.94 & 73.23 & 43.87 & 73.44 & 46.09 & 75.67 & 74.79 & 47.06 & 74.08 & 76.24 & 60.04\\
 & w/ ours & \textbf{39.43} & \textbf{46.96} & \textbf{74.06} & \textbf{43.72} & \textbf{79.62} & \textbf{50.96} & \textbf{75.60} & \textbf{50.12} & \textbf{76.24} & \textbf{50.24} & \textbf{77.87} & \textbf{78.33} & \textbf{52.68} & \textbf{76.40} & \textbf{78.01} & \textbf{63.35}\\
 & $\Delta$ & \textit{+1.99} & \textit{+4.43} & \textit{+1.86} & \textit{+1.35} & \textit{+3.90} & \textit{+5.02} & \textit{+2.37} & \textit{+6.25} & \textit{+2.80} & \textit{+4.15} & \textit{+2.20} & \textit{+3.54} & \textit{+5.62} & \textit{+2.32} & \textit{+1.77} & \textit{+3.30}\\
\bottomrule
\end{tabular}
}
\end{table*}

\begin{table*}[t]
    \caption{
Dice scores of mmFormer, SwinUNETR and their fuzzy-regularized variant on the Pretreat-MetsToBrain-Masks dataset under missing-modality settings. “MN” denotes the number of missing modalities, and “Full” indicates the full-modality setting.
}
    \label{tab:metastasis}
    \centering
    \footnotesize
    \begin{tabular}{c|l|cc|c|cc|cc|cc}
        \toprule
         & & \multicolumn{2}{c|}{All Settings} & Full & \multicolumn{2}{c|}{MN=1} & \multicolumn{2}{c|}{MN=2} & \multicolumn{2}{c}{MN=3} \\
        Type & Method & Mean & Std. & Mean & Mean & Std. & Mean & Std. & Mean & Std. \\
        \midrule
        \multirow{4}{*}{WT} 
        & mmFormer & 42.90 & 14.36 & 66.08 & 54.54 & 10.24 & 41.83 & 9.86 & 27.06 & 9.58 \\
        & w/ ours & 48.78 & 12.40 & \textbf{71.10} & 59.69 & 9.45 & 47.28 & 9.14 & 34.43 & 3.06 \\
        \cmidrule(r){2-11}
        & SwinUNETR & 58.58 & 8.46 & 68.87 & 64.29 & 4.88 & 59.15 & 6.05 & 49.46 & 7.40 \\
        & w/ ours & \textbf{61.74} & 8.42 & 71.03 & \textbf{67.16} & 4.39 & \textbf{62.52} & 5.85 & \textbf{52.84} & 8.27 \\
        \midrule
        \multirow{4}{*}{TC}
        & mmFormer & 40.46 & 14.87 & 64.28 & 52.22 & 10.83 & 39.26 & 10.49 & 24.53 & 10.45 \\
        & w/ ours & 44.76 & 14.61 & 69.23 & 57.54 & 9.88 & 43.86 & 10.01 & 27.21 &5.78 \\
        \cmidrule(r){2-11}
        & SwinUNETR & 54.34 & 16.68 & 69.02 & 62.45 & 13.65 & 54.96 & 15.84 & 41.63 & 15.07 \\
        & w/ ours & \textbf{56.67} & 17.52 & \textbf{72.72} & \textbf{65.35} & 14.25 & \textbf{57.44} & 16.39 & \textbf{42.84} & 15.72 \\
        \midrule
        \multirow{4}{*}{ET}
        & mmFormer & 35.83 & 16.31 & 60.34 & 47.62 & 12.3 & 34.95 & 12.15 & 19.22 & 12.72 \\
        & w/ ours & 38.46 & 16.93 & \textbf{65.80} & 51.84 & 13.08 & 37.60 & 12.09 & 19.51 & 9.72 \\
        \cmidrule(r){2-11}
        & SwinUNETR & 45.16 & 18.66 & 61.71 & 54.24 & 15.91 & 45.39 & 18.32 & 31.60 & 16.15\\
        & w/ ours & \textbf{47.31} & 19.23 & 65.51 & \textbf{56.60} & 15.95 & \textbf{47.68} & 18.66 & \textbf{32.91} & 16.99 \\
        \bottomrule
    \end{tabular}
\end{table*}

\subsection{Results}
\label{sec:results}
We evaluate the robustness of our proposed method using three complementary modality configurations. Initially, the full modality setting is considered, where all MRI sequences are available. This scenario provides a baseline for the model's performance under complete data conditions. 
Then,  the single modality setting is examined, utilizing only one MRI sequence to assess the model's adaptability to data scarcity. Finally, the missing modalities setting is explored, where 1–3 modalities are randomly omittedn to simulate uncertainty in data acquisition.


\begin{figure}[t]
  \centering
  \includegraphics[width=\linewidth]{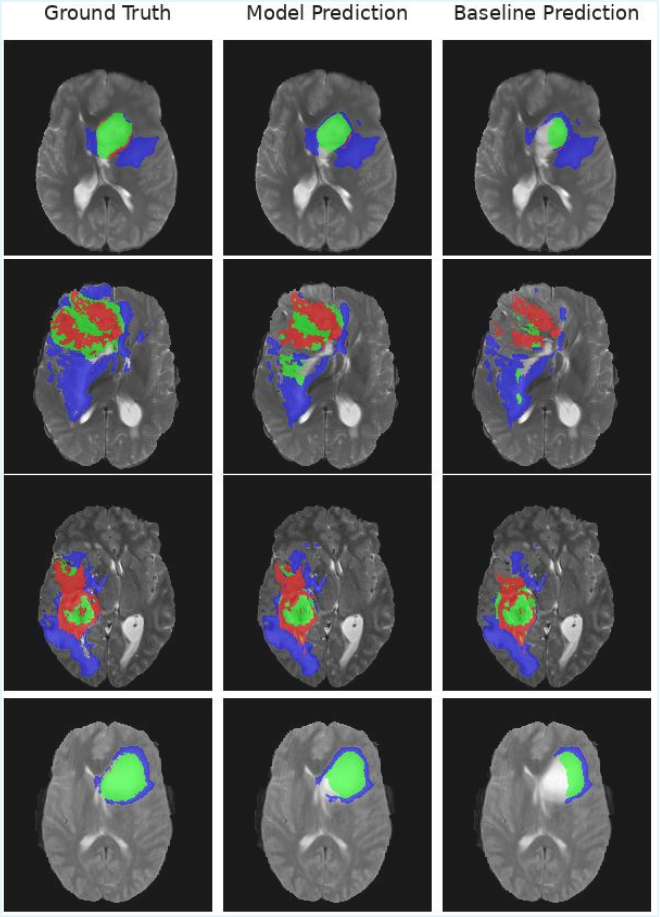}
  \caption{Qualitative visualization of segmentation results on mmformer with BraTS18 dataset. Model Prediction is mmformer + fuzzy, Baseline Prediction is mmformer.}
  \label{fig:visualization}
\end{figure}

\paragraph{Full Modality Setting} \textbf{Our method consistently enhances segmentation accuracy and reliability on the BraTS18 and Pretreat-MetsToBrain-Masks datasets.}
Under the full modality condition, where all four MRI sequences are utilized, results are presented with the mmFormer~\cite{zhang2022mmformer} and SwinUNETR~\cite{hatamizadeh2021swin} baselines in Table~\ref{tab:full_modalities}.  These baselines, integrated with our strategy, show substantial improvements in mean Dice scores and consistent benefits across tumor subregions for both datasets.  As illustrated in Fig.~\ref{fig:visualization}, the baseline models often struggle to delineate large, contiguous lesion areas, especially in regions far from the boundary. In contrast, our proposed method yields smoother and more complete spatial predictions. Baseline models frequently concentrate on high-uncertainty boundary regions, which can lead to overfitting due to ambiguous edge cues, resulting in fragmented and inconsistent segmentations. By incorporating a fuzzy auxiliary loss, our variant maintains coherent structures across the tumor core and enhancing tumor regions, effectively mitigating spurious uncertainty.



\begin{table}
  \caption{Ablation study on the BraTS18 dataset with VNet under the single-modality setting (Dice scores). “fuzzy” denotes the fuzzy-label variant, and “$\rho$” denotes the learnable parameters.}
  \label{tab:Ablation Study}
  \centering
  \footnotesize
  \begin{tabular}{clcccc}
    \toprule
    modality & Setting & mean & wt & tc & et \\
    \midrule
    \multirow{3}{*}{T1}
    & Baseline & 60.23 & 77.60 & 65.02 & 38.07 \\
    & Ours w/o $\rho$ & 61.50 & 78.06 & 67.01 & 39.42\\
    & \textbf{Ours} & 62.09 & 78.49 & 67.74 & 40.03 \\
    \toprule
    \multirow{3}{*}{T2}
    & Baseline& 64.98 & 85.13 & 65.91 & 43.90 \\
    & Ours w/o $\rho$ & 65.93 & 87.16 & 67.65 & 42.98 \\
    & \textbf{Ours} & 66.98 & 87.43 & 70.58 & 42.94 \\
    \toprule
    \multirow{3}{*}{T1ce}
    & Baseline& 75.01 & 76.79 & 81.18 & 67.07 \\
    & Ours w/o $\rho$ & 77.04 & 77.76 & 82.87 & 70.48\\
    & \textbf{Ours} & 77.92 & 79.45 & 82.89 & 71.41 \\
    \toprule
    \multirow{3}{*}{Flair}
    & Baseline& 63.52 & 85.00 & 67.81 & 37.75 \\
    & Ours w/o $\rho$ & 62.68 & 87.21 & 64.78 & 36.05\\
    & \textbf{Ours} & 66.18 & 87.35 & 68.18 & 43.00 \\
    \bottomrule
  \end{tabular}
\end{table}

\paragraph{Single Modality Setting with VNet}
\textbf{Our method improves segmentation accuracy, even with limited modality information.}
In Table~\ref{tab:Ablation Study}, the first row of each block presents the baseline results, while the last row shows the outcomes after incorporating our strategy. Both sets of results pertain to the BraTS18 dataset, using only one MRI modality input with the VNet architecture~\cite{milletari2016v}.
Even under this extreme configuration, our approach consistently surpasses the baseline in mean Dice scores across most tumor subregions across all four modalities.
This demonstrates  the effectiveness of our Region-wise Curriculum Learning Strategy in enhancing discriminative region learning when modality-specific information is limited. Instead of solely refining boundaries, our adaptive fuzzy constraint stabilizes prediction confidence in ambiguous regions, leading to robust optimization and reduced sensitivity to local uncertainties.


\paragraph{Missing-Modality Setting}
\textbf{Our strategy consistently enhances performance stability across various missing-modality configurations.
}
Tables~\ref{tab:swinunetr} and \ref{tab:metastasis} provide quantitative results for the BraTS18 and Pretreat-MetsToBrain-Masks datasets when one or more input modalities are absent, evaluated using U-HeMIS~\cite{havaei2016hemis}, U-HVED~\cite{dorent2019hetero}, mmFormer~\cite{zhang2022mmformer}, and SwinUNETR~\cite{hatamizadeh2021swin} architectures. The results for U-HeMIS, U-HVED, and mmFormer are sourced from \cite{zhang2022mmformer}. Since SwinUNETR does not report results on the BraTS18 dataset, we conducted these experiments ourselves. Due to time limitations, both the baseline and mmFormer with our strategy were trained for 300 epochs on the Pretreat-MetsToBrain-Masks dataset, with extended 1000-epoch results available in the supplementary material.


Across a range of configurations, from a single remaining modality to the random removal of up to three sequences, integrating our fuzzy auxiliary loss generally results in performance gains.  Average Dice score improvements range from $+0.79$ to $+3.67$ points in the single modality case and remain stable through various multi-missing configurations, demonstrating notable robustness to incomplete input data. 
The most significant improvements are observed in the \texttt{ET} (enhancing tumor) subregion, where cross modality information is most critical, suggesting that  the proposed loss's capacity to compensate for uncertainties due to missing contextual information by promoting coherent region-level consistency.
mmFormer utilizes multiple loss branches, while our approach is applied only to the fused branch. This limited integration might restrict its benefits on the \texttt{ET}, resulting in not superior performance to the baseline.
Our approach significantly improves robustness against modality dropout, showcasing its potential for real-world clinical scenarios where MRI sequences are incomplete.

\begin{figure}[t]
  \centering
  \includegraphics[width=\linewidth]{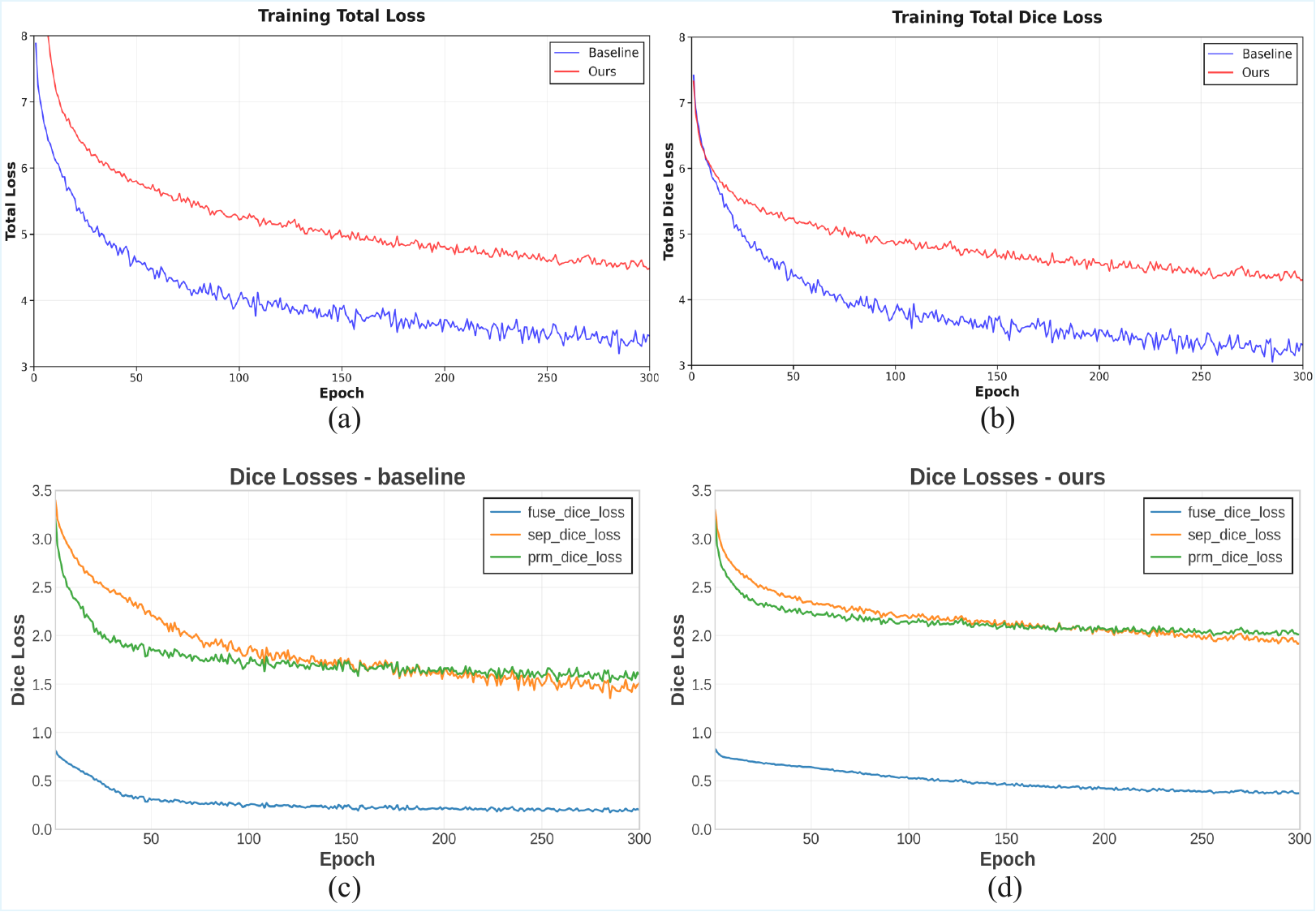}
  \caption{Training Stability Analysis on mmFormer under the missing modalities}
  \label{fig:stability}
\end{figure}

\subsection{Training Stability Analysis}
To further examine the training behavior, we visualize the loss curves of mmFormer under the missing-modality setting in Fig~\ref{fig:stability}, where (a) shows the total loss and (b) presents the Dice loss.
Both curves demonstrate that our method achieves smoother convergence, while the baseline exhibits larger oscillations during training.
Fig~\ref{fig:stability}(c) and (d) illustrate the per-branch Dice Loss results of the baseline and our method, respectively.
These observations confirm that our approach effectively suppresses unstable gradient fluctuations and leads to a more reliable training process.

\subsection{Ablation Study}

To assess the effect of each proposed component, we conduct an ablation study on the brain metastasis and non-metastatic tumor segmentation task using VNet under single modality settings. As shown in Table~\ref{tab:Ablation Study}, we compare three variants: (1) Baseline, the standard VNet based architecture without uncertainty modeling; (2) Ours w/o $\rho$, which introduces the proposed IFL representation to isolate high uncertainty regions but without adaptive parameterization; and (3) Ours, the full version incorporating the learnable parameters that enable adaptive Pareto consistent refinement during training.
Across all modalities, the uncertainty-aware variants outperform the baseline, confirming the benefit of decoupling high and low uncertainty regions in the early training stage. The version with learnable parameters ($\rho$) further improves performance, where data uncertainty and annotation variability are higher. This indicates that the adaptive parameterization effectively refines the balance between different uncertainty regions, guiding the optimization toward more stable and globally consistent segmentation. In particular, the improvements are consistent across all modalities, highlighting the robustness of the proposed uncertainty guided optimization framework.

\section{Conclusion}
This study introduces a novel \textit{Region-wise Curriculum Learning Strategy} designed to address the inherent non-uniformity of uncertainty in medical image segmentation, with a particular focus on boundary regions where ambiguity is most pronounced.  The proposed Intuitionistic Fuzzy Label representation enhances boundary region, which allows for smooth label transitions across uncertain boundaries while preserving binary confidence in stable regions. Then, by prioritizing the learning process from regions of low uncertainty and gradually incorporating more uncertain boundary areas, our approach stabilizes optimization and reduces gradient variance, facilitating more reliable convergence toward Pareto-approximate solutions. The proposed Pareto-consistent formulation effectively balances the trade-offs between different regional uncertainties, reshaping the loss landscape to guide the model toward improved segmentation accuracy. Extensive evaluations conducted on two prominent medical segmentation benchmarks demonstrate that our approach significantly enhances both segmentation performance and training stability compared to traditional crisp-set baselines. Notably, these improvements are evident across a variety of tasks, including conventional segmentation, scenarios with missing modalities, and single-modality challenges. This highlights the robustness and adaptability of our method in handling diverse medical imaging tasks.

\paragraph{Limitations and future work}
Although mmFormer employs multiple loss branches, our current implementation introduces our strategy only to the fused branch. This design choice may limit its impact on certain subregions (e.g., \texttt{ET}). In future work, we plan to extend our strategy to additional branches.



%% file: sec/X_suppl.tex
\clearpage
\setcounter{page}{1}
\maketitlesupplementary

\section{Fuzzy Regularization Smooths the Loss Landscape: A Mathematical Explanation}

\subsection{Definition}

For a pixel and a given class $c$, let $p_c \in (0,1)$ denote the predicted probability and  
$\mu_c \in [0,1]$ the fuzzy membership degree.  
The proposed fuzzy regularization term is
\begin{equation}
\mathcal{L}_{\text{fuzzy}}
= - \mu_c \log p_c
  - \rho_2 (1-\mu_c)\, \log\big(\rho_1(1-p_c)\big),
\label{eq:fuzzy-def}
\end{equation}
where $\rho_1, \rho_2 \in (0,1]$ are learnable parameters ensuring boundedness of the non-membership term.

Using the log identity
\[
\log(\rho_1(1-p_c)) = \log \rho_1 + \log(1-p_c),
\]
the constant $\log\rho_1$ plays no role when differentiating w.r.t.~$p_c$.

\subsection{First-Order Derivatives}

Differentiating Eq.~\eqref{eq:fuzzy-def} with respect to $p_c$ gives
\begin{equation}
\frac{\partial \mathcal{L}_{\text{fuzzy}}}{\partial p_c}
= - \frac{\mu_c}{p_c}
  + \frac{\rho_2(1-\mu_c)}{1-p_c}.
\label{eq:first-deriv}
\end{equation}

For comparison, the binary cross-entropy (CE)
\[
\mathcal{L}_{\text{CE}}
= - y_c \log p_c - (1-y_c)\log(1-p_c)
\]
satisfies
\begin{equation}
\frac{\partial \mathcal{L}_{\text{CE}}}{\partial p_c}
= - \frac{y_c}{p_c}
  + \frac{1-y_c}{1-p_c}.
\label{eq:first-deriv-ce}
\end{equation}

\subsection{Second-Order Derivatives}

Differentiating Eq.~\eqref{eq:first-deriv} yields the curvature:
\begin{equation}
\frac{\partial^2 \mathcal{L}_{\text{fuzzy}}}{\partial p_c^2}
= \frac{\mu_c}{p_c^2}
  + \frac{\rho_2(1-\mu_c)}{(1-p_c)^2}.
\label{eq:second-deriv}
\end{equation}

CE has curvature
\begin{equation}
\frac{\partial^2 \mathcal{L}_{\text{CE}}}{\partial p_c^2}
= \frac{y_c}{p_c^2}
  + \frac{1-y_c}{(1-p_c)^2}.
\label{eq:second-deriv-ce}
\end{equation}

Both are strictly non-negative, hence both losses are convex w.r.t.~$p_c$ in this 1D projection.

\subsection{Why Fuzzy Regularization Smooths the Loss Landscape}
\label{landscape}

The smoothing effect does not come from fuzzy loss having smaller curvature at every point, but from shifting the optimization target away from the sharp boundaries \(p_c\in\{0,1\}\).

Under hard labels, CE enforces \(p_c \rightarrow y_c \in \{0,1\}\). Near these boundaries, the curvature
\[
\frac{\partial^2 \mathcal{L}_{\mathrm{CE}}}{\partial p_c^2}
= \frac{y_c}{p_c^2} + \frac{1-y_c}{(1-p_c)^2}
\]
can grow arbitrarily large when \(p_c\) approaches \(0\) or \(1\), yielding steep and unstable regions in the loss landscape.

With fuzzy labels \(\mu_c\in(0,1)\), the minimizer of the fuzzy loss satisfies \(p_c^\star \approx \mu_c\), which lies strictly inside \((0,1)\). Evaluating the curvature at this equilibrium,
\[
\left.\frac{\partial^2 \mathcal{L}_{\mathrm{fuzzy}}}{\partial p_c^2}\right|_{p_c = p_c^\star}
= \frac{\mu_c}{(p_c^\star)^2}
+ \frac{\rho_2(1-\mu_c)}{(1-p_c^\star)^2},
\]
remains finite because both \(p_c^\star\) and \(1-p_c^\star\) are bounded away from zero.

Fuzzy regularization does not reduce curvature everywhere; instead, it prevents the optimizer from entering the high-curvature boundary regions enforced by hard labels. By shifting the optimum to an interior point \(p_c^\star\in(0,1)\), it keeps curvature bounded along training trajectories, resulting in a smoother and more stable optimization landscape.

\subsection{Landscape Smoothing Effect}

Let $\theta$ denote model parameters, and let $p_c = p_c(\theta)$.  
The Hessian of the full loss involves
\[
\nabla_\theta^2 \mathcal{L}
= \frac{\partial^2 \mathcal{L}}{\partial p_c^2}
(\nabla_\theta p_c)(\nabla_\theta p_c)^\top
+ \frac{\partial \mathcal{L}}{\partial p_c}\,\nabla_\theta^2 p_c.
\]

Since the architecture term $\nabla_\theta^2 p_c$ is fixed for both losses, the only difference arises from the scalar curvature term  
\(\partial^2 \mathcal{L}/\partial p_c^2\).  
As shown in Sec.~\ref{landscape}, the fuzzy loss avoids the exploding curvature regimes associated with hard labels at uncertain pixels. By bounding the scalar curvature term $\partial^2 \mathcal{L}/\partial p_c^2$, it reduces:
\begin{itemize}
    \item the maximal Hessian eigenvalue,
    \item the condition number of the loss surface,
    \item the prevalence of sharp local minima.
\end{itemize}

As a consequence, optimization trajectories under SGD or Adam experience a flatter, more stable loss landscape, especially around anatomically ambiguous boundaries.

\subsection{Conclusion of This Section}

Fuzzy regularization preserves per-pixel convexity while avoiding the extremely high-curvature regions induced by hard labels at ambiguous pixels.
This attenuates the sharp Hessian directions induced by hard, unreliable labels, yielding a smoother and more optimizable landscape.  
Empirically, this manifests as lower variance in training loss curves and improved stability under missing-modality conditions.

\begin{figure}[t]
    \centering
    \includegraphics[width=0.95\linewidth]{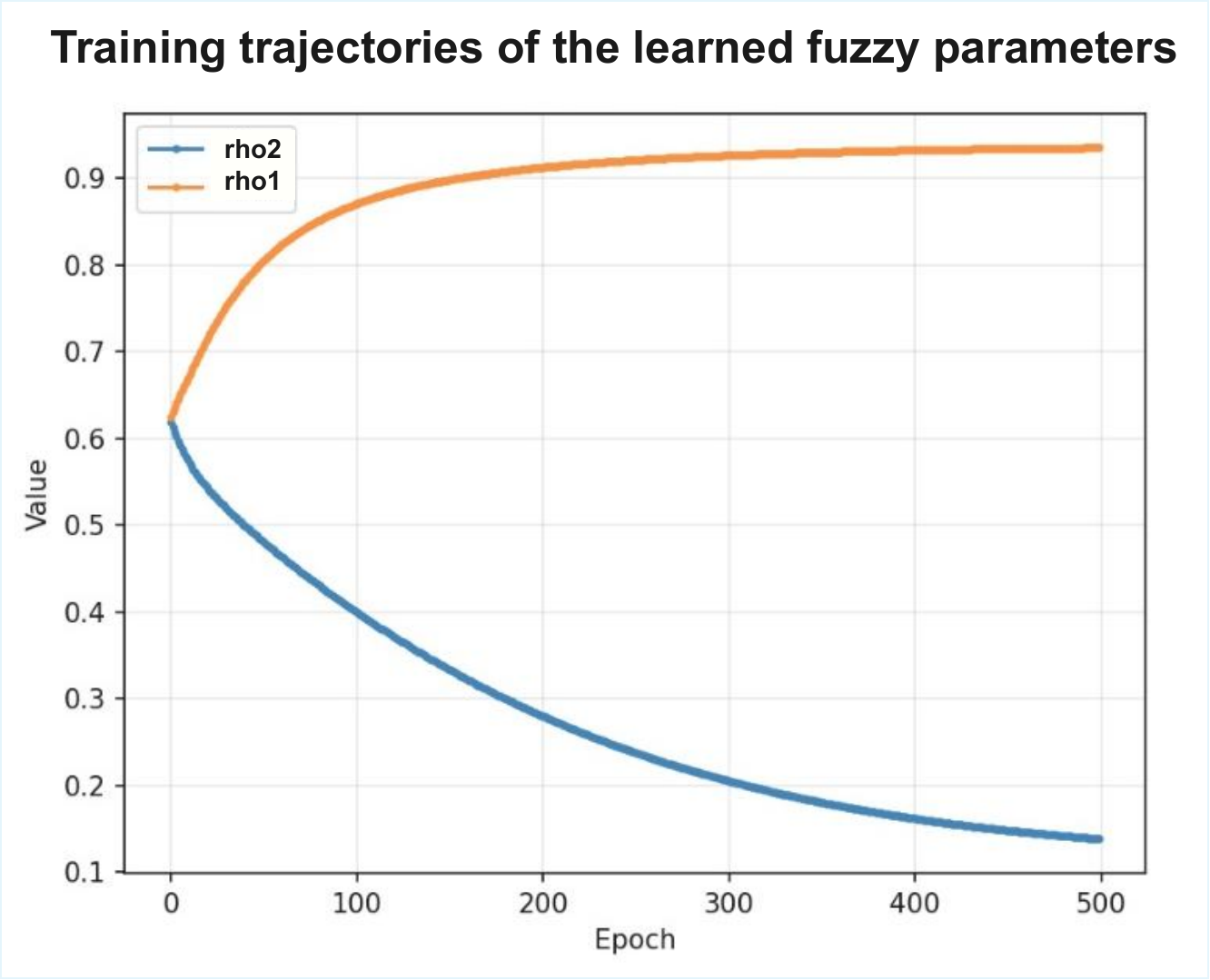}
    \caption{
        Evolution of the learnable fuzzy parameters $\rho_1$ and $\rho_2$ during training.
        Consistent with the gradient analysis in Sec.~\ref{sec:rho_analysis}, $\rho_1$ monotonically increases towards $1^-$, indicating growing confidence in the logits-based membership, while $\rho_2$ asymptotically decays towards $0^+$, gradually annealing the regularization from the non-membership term.
    }
    \label{fig:rho}
\end{figure}

\section{Analysis of the Learned Parameters $\rho_1$ and $\rho_2$}
\label{sec:rho_analysis}

To understand the empirical behavior of the learnable parameters $\rho_1$ and $\rho_2$ observed in Figure~\ref{fig:rho}, we analyze the gradients of the fuzzy auxiliary loss defined in Eq.~(\ref{eq:fuzzy_loss}).
Recall the loss formulation for a single pixel and class $c$:
\begin{equation}
\mathcal{L}_{\text{fuzzy}} = - \mu_c \log p_c - \rho_2 (1-\mu_c) \log (\rho_1 (1 - p_c)).
\end{equation}
Here, $\rho_1, \rho_2 \in (0,1)$ are parameterized via a sigmoid function. The dynamics of these parameters are coupled and evolve through different training phases.

\subsection{Coupled Dynamics and Asymptotic Behavior}

\paragraph{Gradient of $\rho_1$:}
Differentiating with respect to $\rho_1$:
\begin{equation}
\frac{\partial \mathcal{L}_{\text{fuzzy}}}{\partial \rho_1} 
= - \frac{\rho_2 (1-\mu_c)}{\rho_1}.
\end{equation}
Assuming $\rho_2 > 0$ and the existence of uncertain regions where $\mu_c < 1$, this gradient is strictly negative. This provides a consistent driving force for $\rho_1$ to increase. Although the magnitude of this force diminishes as $\rho_2$ decays, $\rho_1$ will monotonically grow towards its upper bound as long as $\rho_2$ remains non-zero. Thus:
\begin{equation}
\rho_1 \rightarrow 1^-.
\end{equation}

\paragraph{Gradient of $\rho_2$:}
Differentiating with respect to $\rho_2$:
\begin{equation}
\frac{\partial \mathcal{L}_{\text{fuzzy}}}{\partial \rho_2} 
= - (1-\mu_c) \log (\rho_1 (1 - p_c)).
\end{equation}
The sign of this gradient depends on the term $\log(\rho_1(1-p_c))$.
\begin{itemize}
    \item \textbf{Early Training Phase:} Initially, predictions are uncertain ($p_c$ is low) and $\rho_1$ is initialized at an intermediate value (e.g., 0.5). The term $\rho_1(1-p_c)$ is typically well below 1, making the log term negative and the gradient positive. This drives $\rho_2$ to decrease.
    \item \textbf{Late Training Phase:} As the model converges, for correct predictions, $p_c \to 1$, ensuring $\log(\rho_1(1-p_c)) \ll 0$. This maintains a positive gradient, further pushing $\rho_2$ towards 0.
    \item \textbf{Boundary Condition:} In the rare case of severe misclassification where $p_c \approx 0$ and $\rho_1 \approx 1$, the term $\rho_1(1-p_c)$ could approach 1, potentially stalling the decay. However, the dominant contribution from correctly classified and uncertain pixels (where $1-p_c < 1$) ensures that the aggregate gradient over the dataset remains positive.
\end{itemize}
Consequently, $\rho_2$ is driven towards its lower bound:
\begin{equation}
\rho_2 \rightarrow 0^+.
\end{equation}

\subsection{Interpretation}
The coupled evolution of $\rho_1$ and $\rho_2$ implements an implicit curriculum:
\begin{itemize}
    \item $\rho_1 \to 1^-$: The model progressively learns to trust the logits-based membership fully, removing the scaling attenuation on the non-target probability $(1-p_c)$.
    \item $\rho_2 \to 0^+$: The regularization weight anneals over time. The decay of $\rho_2$ is naturally slower than the rise of $\rho_1$ (as $\rho_2$'s gradient depends on the log magnitude), allowing the fuzzy regularization to guide the optimization during the critical early phases before fading out to let the main Dice loss dominate precision refinement.
\end{itemize}

\begin{figure}[t]
  \centering
  \includegraphics[width=\linewidth]{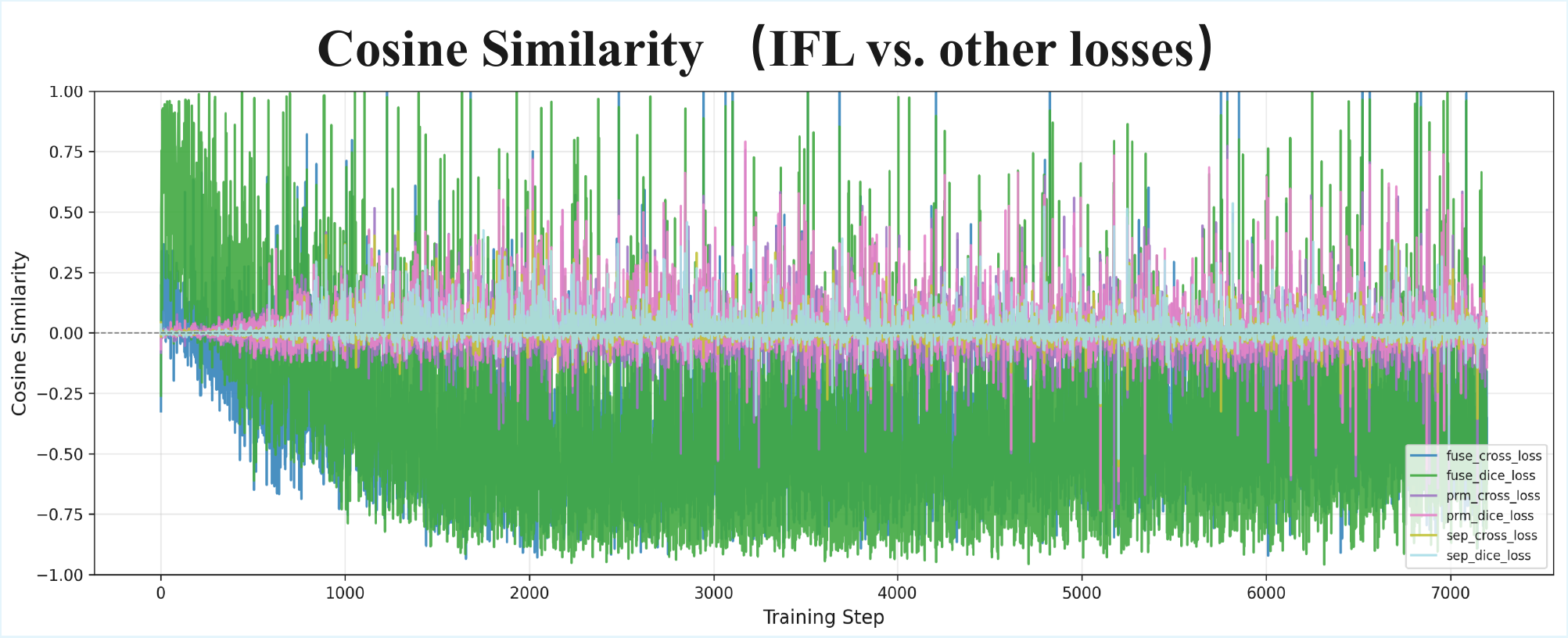}
  \caption{Cosine similarity between our IFL and other losses during training.}
  \label{fig:loss_conflict}
\end{figure}

\section{Gradient Interaction Analysis.}
We analyze the interaction between the proposed IFL and other training objectives by monitoring the cosine similarity between their gradients throughout training (Fig.~\ref{fig:loss_conflict}). A time-varying pattern is observed on the fuse branch where IFL is applied. In the early stage, IFL shows positive cosine similarity with the fuse losses, indicating cooperative optimization directions. As training progresses, the cosine values gradually become negative, suggesting that IFL and the fuse losses start to update the shared parameters in competing directions.

In contrast, the cosine similarity between IFL and the other decoder losses remains close to zero across the entire optimization process. This implies that IFL primarily influences the parameter subspace coupled to the fuse branch, while its effective gradient components projected onto other decoder heads remain small.

This result shows that IFL induces a localized and time-dependent gradient interaction: cooperative at the beginning to stabilize learning, and mildly competitive in later stages when enforcing boundary-related corrections, which may effect on ET.

\begin{table*}[t]
\caption{
This table provides the extended 1000-epoch comparison between mmFormer and mmFormer w/ ours on the Pretreat-MetsToBrain-Masks dataset under missing-modality settings, supplementing the 300-epoch results discussed previously in table 3.
Here, WT, TC, and ET denote Whole Tumor, Tumor Core, and Enhancing Tumor, respectively.
The modality labels T1, T2, T1ce, and F correspond to T1, T2, T1ce, and FLAIR, respectively,
Full indicates the full-modality setting.
}
\label{tab:mmformer_ablation}
\centering
\resizebox{\textwidth}{!}{
\begin{tabular}{c|l|cccc|cccccc|cccc|c|c}
\toprule
 &  & \multicolumn{4}{c|}{Remain 1 modality} & \multicolumn{6}{c|}{Remain 2 modalities} & \multicolumn{4}{c|}{Missing 1 modality} & Full & Avg. \\
\midrule
Type & Method & T1 & T2 & T1ce & FLAIR & T1,T2 & T1,T1ce & T1,F & T2,T1ce & T2,F & T1ce,F & T1 & T2 & T1ce & FLAIR &  &  \\
\midrule

\multirow{2}{*}{WT}
& mmFormer &
58.22 & 42.57 & 49.78 & 55.71 &
58.83 & 62.49 & 65.65 & 50.50 & 57.20 & 57.28 &
58.02 & 65.81 & 65.51 & 62.03 &
65.92 & 58.37 \\

& mmFormer w/ ours &
\textbf{62.55} & \textbf{44.90} & \textbf{52.70} & \textbf{60.39} &
\textbf{65.33} & \textbf{66.92} & \textbf{73.16} & \textbf{55.47} & \textbf{62.33} & \textbf{62.06} &
\textbf{63.06} & \textbf{72.71} & \textbf{73.15} & \textbf{68.08} &
\textbf{72.52} & \textbf{63.69} \\
\midrule

\multirow{2}{*}{TC}
& mmFormer &
54.92 & 40.61 & 47.42 & 53.32 &
56.47 & 60.39 & 63.65 & 47.98 & 54.82 & 55.25 &
55.72 & 63.92 & 63.53 & 59.87 &
63.97 & 56.19 \\

& mmFormer w/ ours &
\textbf{60.50} & \textbf{42.27} & \textbf{50.27} & \textbf{58.04} &
\textbf{63.33} & \textbf{65.05} & \textbf{71.36} & \textbf{53.02} & \textbf{60.17} & \textbf{59.86} &
\textbf{60.85} & \textbf{70.89} & \textbf{71.35} & \textbf{66.23} &
\textbf{70.75} & \textbf{61.60} \\
\midrule

\multirow{2}{*}{ET}
& mmFormer &
48.82 & \textbf{37.96} & 45.86 & 51.68 &
48.70 & 52.90 & 58.95 & 46.01 & 52.42 & 53.09 &
52.47 & 59.65 & 57.19 & 54.51 &
58.18 & 51.88 \\

& mmFormer w/ ours &
\textbf{52.58} & \textbf{37.96} & \textbf{45.99} & \textbf{52.78} &
\textbf{55.66} & \textbf{57.35} & \textbf{64.30} & \textbf{47.78} & \textbf{56.60} & \textbf{56.48} &
\textbf{57.18} & \textbf{63.89} & \textbf{64.95} & \textbf{58.61} &
\textbf{64.30} & \textbf{55.76} \\
\bottomrule
\end{tabular}
}
\end{table*}

\begin{table*}[t]
\caption{
Full version of Table 3.
Here, WT, TC, and ET denote Whole Tumor, Tumor Core, and Enhancing Tumor, respectively.
The modality labels T1, T2, T1ce, and F correspond to T1, T2, T1ce, and FLAIR, respectively, and Full indicates the full-modality setting.
}
\label{tab:swinunetr}
\centering
\resizebox{\textwidth}{!}{
\begin{tabular}{c|l|cccc|cccccc|cccc|c|c}
\toprule
 &  & \multicolumn{4}{c|}{Remain 1 modality} & \multicolumn{6}{c|}{Missing 2 modalities} & \multicolumn{4}{c|}{Missing 1 modality} & Full & Avg. \\
\midrule
 Type & Method & T1 & T2 & T1ce & FLAIR & T1,T2 & T1,T1c & T1,F & T2,T1c & T2,F & T1c,F & T1 & T2 & T1ce & FLAIR &  &  \\
\midrule
\multirow{4}{*}{WT}
& mmFormer & 12.67 & 29.9 & 39.38 & 26.32 & 48.61 & 27.93 & 50.68 & 31.27 & 49.31 & 43.21 & 53.93 & 63.33 & 40.36 & 60.55 & 66.08 & 42.90 \\
& mmFormer w/ ours & 32.85 & 37.79 & 37.31 & 30.18 & 44.08 & 36.00 & 51.40 & 38.56 & 58.38 & \textbf{55.30} & 51.70 & 68.49 & 51.35 & \textbf{67.25} & \textbf{71.10} & 43.78 \\
\cmidrule(r){2-18}
& SwinUNETR & 38.41 & 47.03 & 56.17 & 56.26 & 67.75 & 60.44 & 61.81 & 58.72 & 56.75 & 49.41 & 69.41 & 67.51 & 59.86 & 60.38 & 68.87 & 58.58 \\
& SwinUNETR w/ ours & \textbf{40.96} & \textbf{49.25} & \textbf{60.88} & \textbf{60.28} & \textbf{70.20} & \textbf{61.57} & \textbf{67.43} & \textbf{60.58} & \textbf{61.92} & 53.39 & \textbf{70.95} & \textbf{70.17} & \textbf{61.36} & 66.18 & 71.03 & \textbf{61.74} \\
\midrule
\multirow{4}{*}{TC}
& mmFormer & 9.15 & 27.72 & 38.24 & 23.02 & 45.47 & 24.84 & 48.58 & 27.4 & 47.75 & 41.52 & 51.40 & 61.33 & 37.26 & 58.9 & 64.28 & 40.45 \\
& mmFormer w/ ours & 26.20 & 30.71 & 33.64 & 18.32 & 41.38 & 32.60 & 46.11 & 33.29 & 56.88 & \textbf{52.93} & 49.74 & 66.48 & \textbf{48.26} & 65.69 & 69.23 & 44.76 \\
\cmidrule(r){2-18}
& SwinUNETR & \textbf{29.25} & 32.91 & 67.3 & \textbf{37.09} & 70.31 & 43.21 & 68.79 & 42.90 & 68.58 & 35.99 & 69.06 & 69.96 & 41.98 & 68.80 & 69.02 & 54.34 \\
& SwinUNETR w/ ours & 28.45 & \textbf{36.71} & \textbf{69.44} & 36.76 & \textbf{72.62} & \textbf{44.18} & \textbf{73.96} & \textbf{42.74} & \textbf{70.43} & 40.72 & \textbf{72.65} & \textbf{71.91} & 43.97 & \textbf{72.87} & \textbf{72.72} & \textbf{56.67} \\
\midrule
\multirow{4}{*}{ET}
& mmFormer & 6.94 & 26.82 & 36.20 & 6.95 & 42.73 & 20.59 & 44.45 & 18.56 & 45.06 & 38.33 & 47.64 & 57.89 & 30.33 & 54.65 & 60.34 & 35.83 \\
& mmFormer w/ ours & 13.87 & \textbf{27.01} & 30.60 & 6.57 & 41.56 & 26.84 & 43.27 & 19.37 & 52.17 & \textbf{42.42} & 49.43 & 63.13 & \textbf{34.36} & 60.47 & \textbf{65.80} & 38.45 \\
\cmidrule(r){2-18}
& SwinUNETR & \textbf{18.81} & 23.33 & 59.32 & 24.97 & 63.21 & 31.23 & 61.71 & 29.92 & 61.1 & 25.16 & 62.96 & 62.29 & 30.39 & 61.33 & 61.71 & 45.16 \\
& SwinUNETR w/ ours & 18.49 & 25.30 & \textbf{61.91} & \textbf{25.96} & \textbf{64.96} & \textbf{32.49} & \textbf{65.74} & \textbf{31.48} & \textbf{63.26} & 28.17 & \textbf{64.61} & \textbf{64.11} & 32.68 & \textbf{65.00} & 65.51 &  \textbf{47.31} \\
\bottomrule
\end{tabular}
}
\end{table*}

\begin{figure*}[t]
  \centering
  \includegraphics[width=\linewidth]{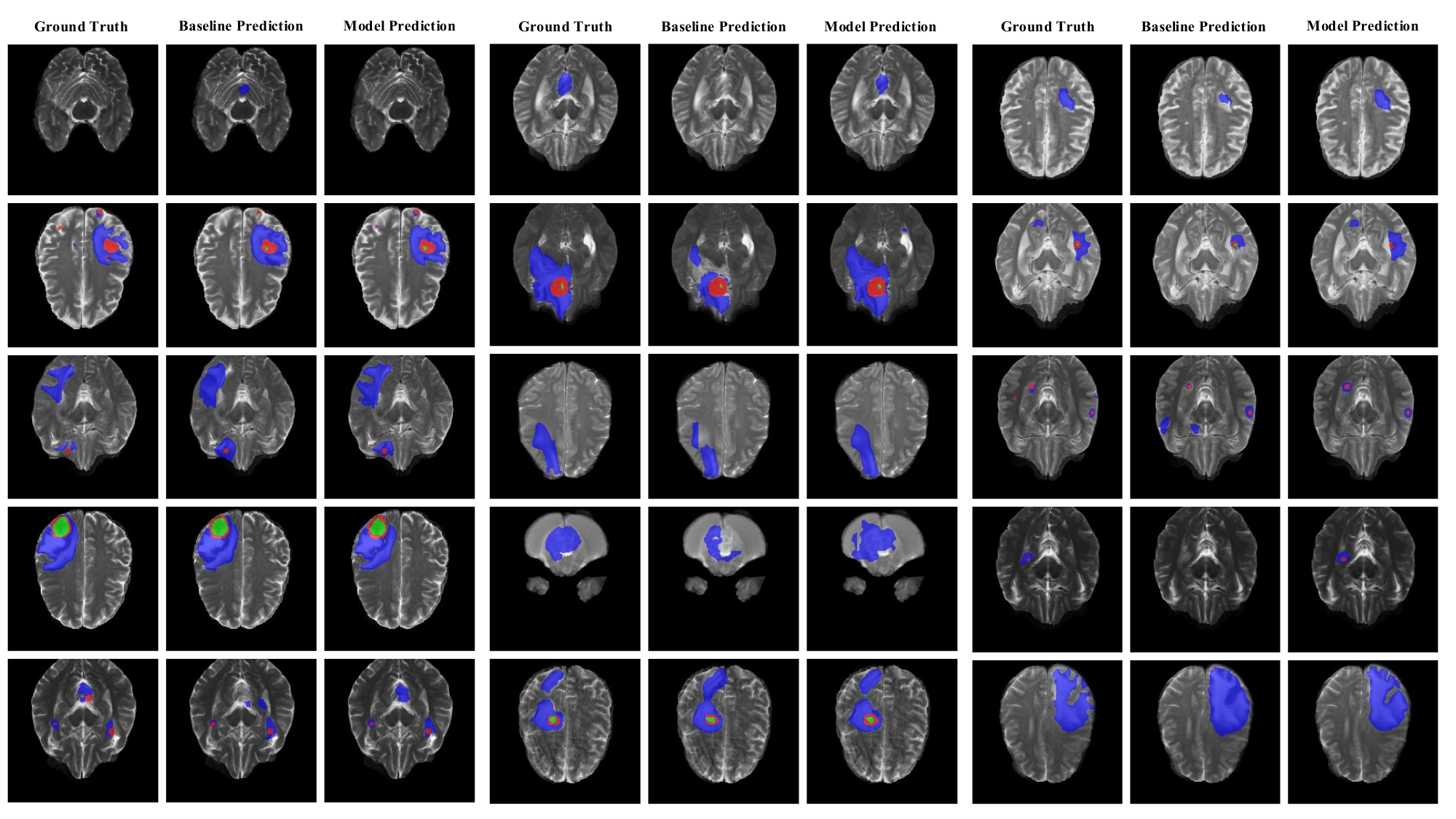}
  \caption{Qualitative visualization of segmentation results produced by mmFormer on the Pretreat-MetsToBrain-Masks dataset. Model Prediction is mmFormer + fuzzy, Baseline Prediction is mmFormer.}
  \label{fig:swinunetr_metastasis}
\end{figure*}

\begin{figure*}[t]
  \centering
  \includegraphics[width=\linewidth]{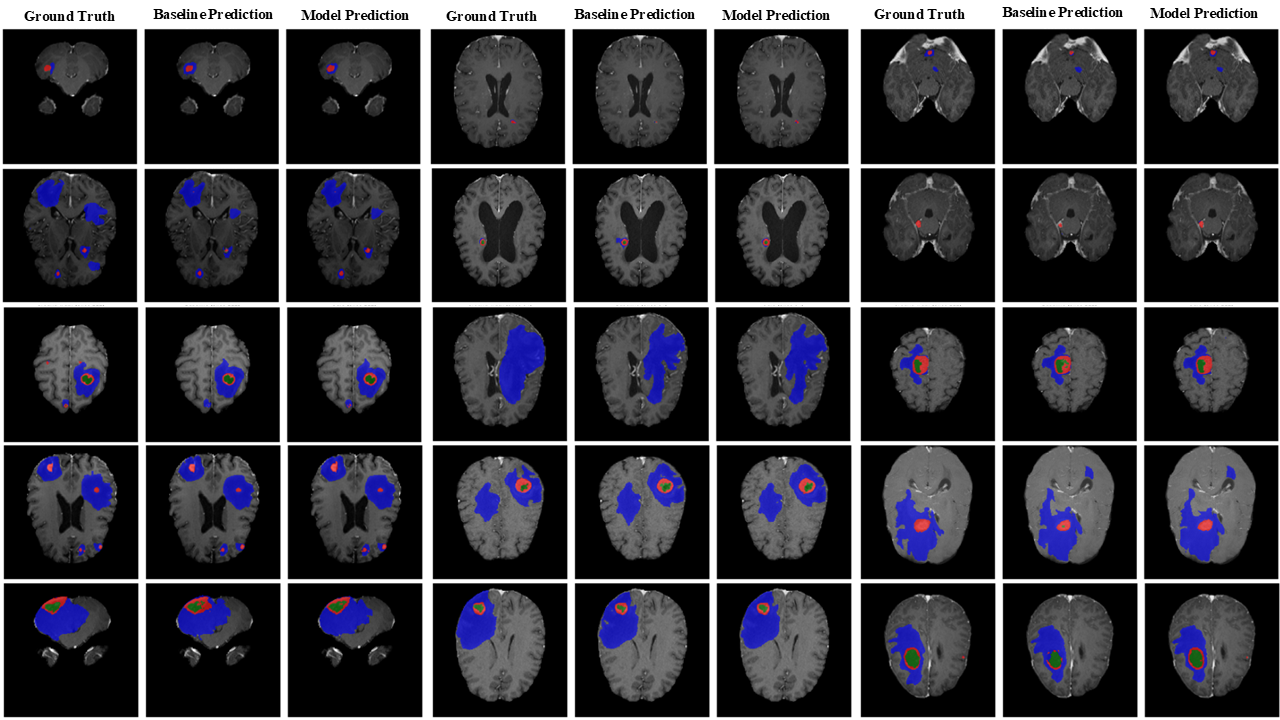}
  \caption{Qualitative visualization of segmentation results produced by SwinUNETR on the Pretreat-MetsToBrain-Masks dataset. Model Prediction is SwinUNETR + fuzzy, Baseline Prediction is SwinUNETR.}
  \label{fig:swinunetr_metastasis}
\end{figure*}

\begin{figure*}[t]
  \centering
  \includegraphics[width=\linewidth]{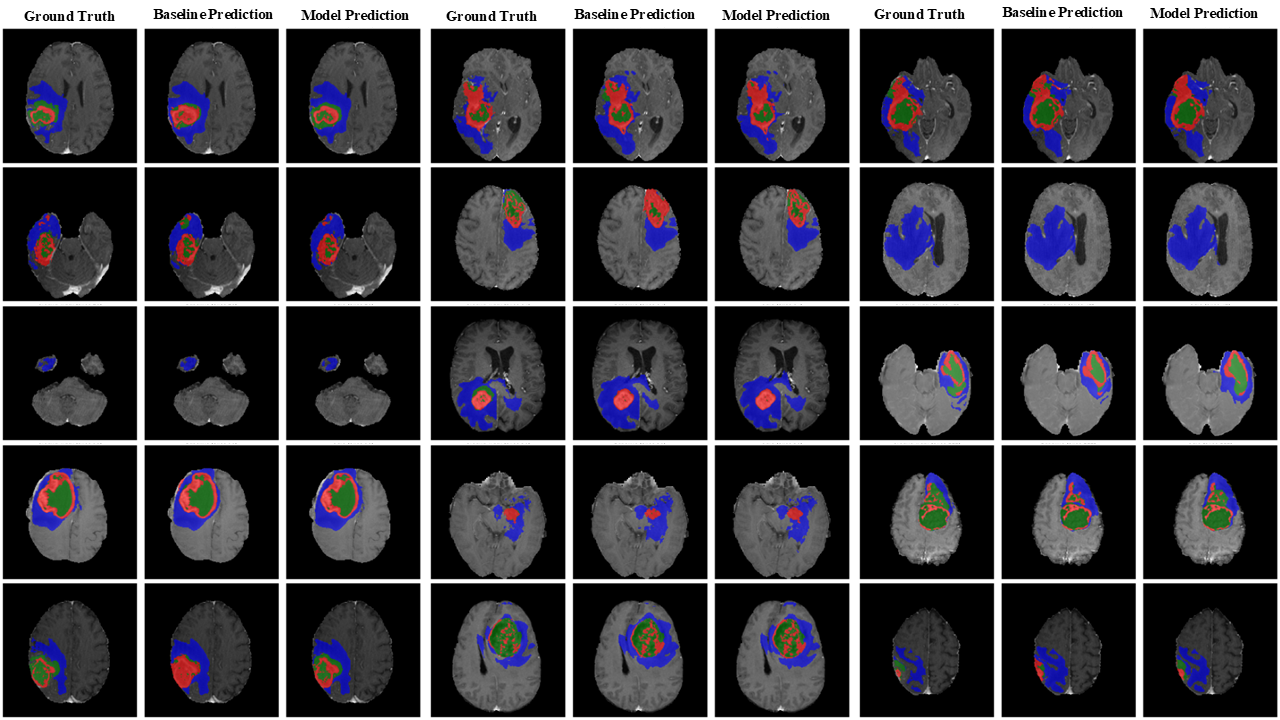}
  \caption{Qualitative visualization of segmentation results produced by SwinUNETR on the BraTS18 dataset. Model Prediction is SwinUNETR + fuzzy, Baseline Prediction is SwinUNETR.}
  \label{fig:swinunetr_Brats1}
\end{figure*}

\begin{figure*}[t]
  \centering
  \includegraphics[width=\linewidth]{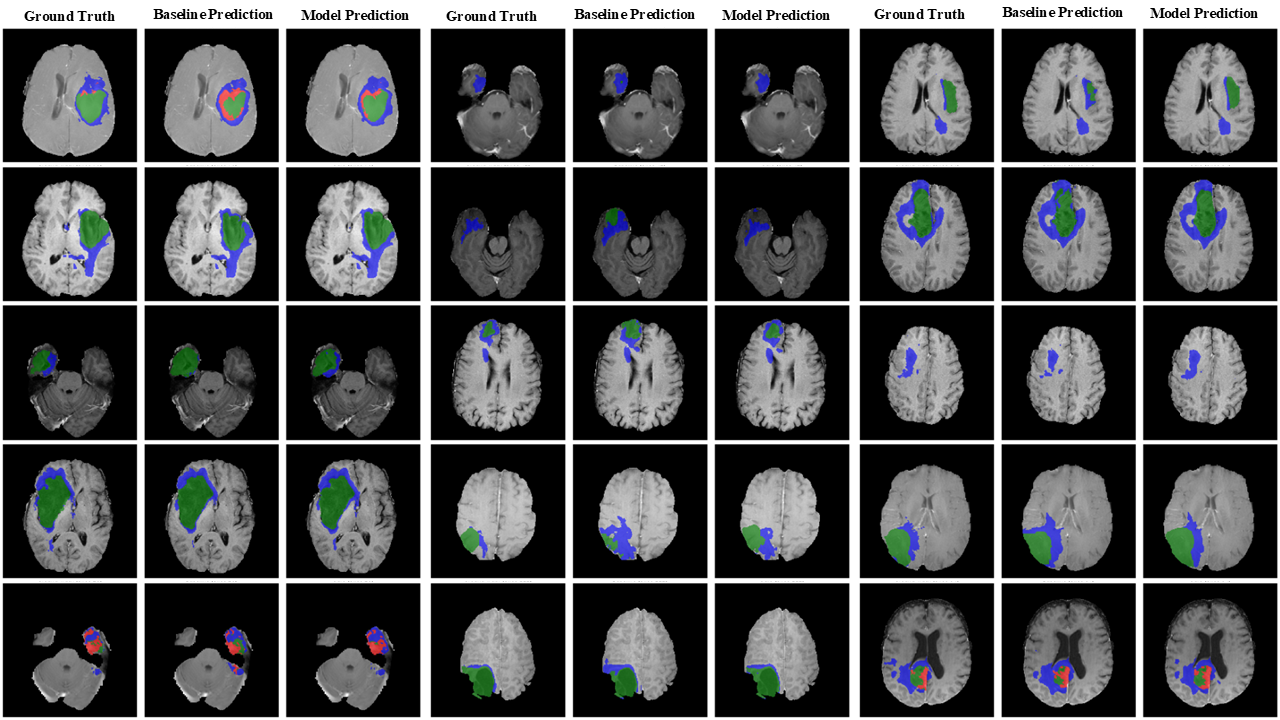}
  \caption{Qualitative visualization of segmentation results produced by SwinUNETR on the BraTS18 dataset. Model Prediction is SwinUNETR + fuzzy, Baseline Prediction is SwinUNETR.}
  \label{fig:swinunetr_Brats1}
\end{figure*}

\section{Complete 1000-Epoch mmFormer Training}
To validate that the performance improvement is not due to insufficient training, we additionally train both the baseline and our fuzzy variant for 1000 epochs on the Pretreat-MetsToBrain-Masks dataset. The training protocol and hyperparameters remain identical to those in the main paper. The quantitative results in Table~\ref{tab:mmformer_ablation} show that our method continues to outperform the baseline under extended training.

\begin{figure*}[t]
  \centering
  \includegraphics[width=\linewidth]{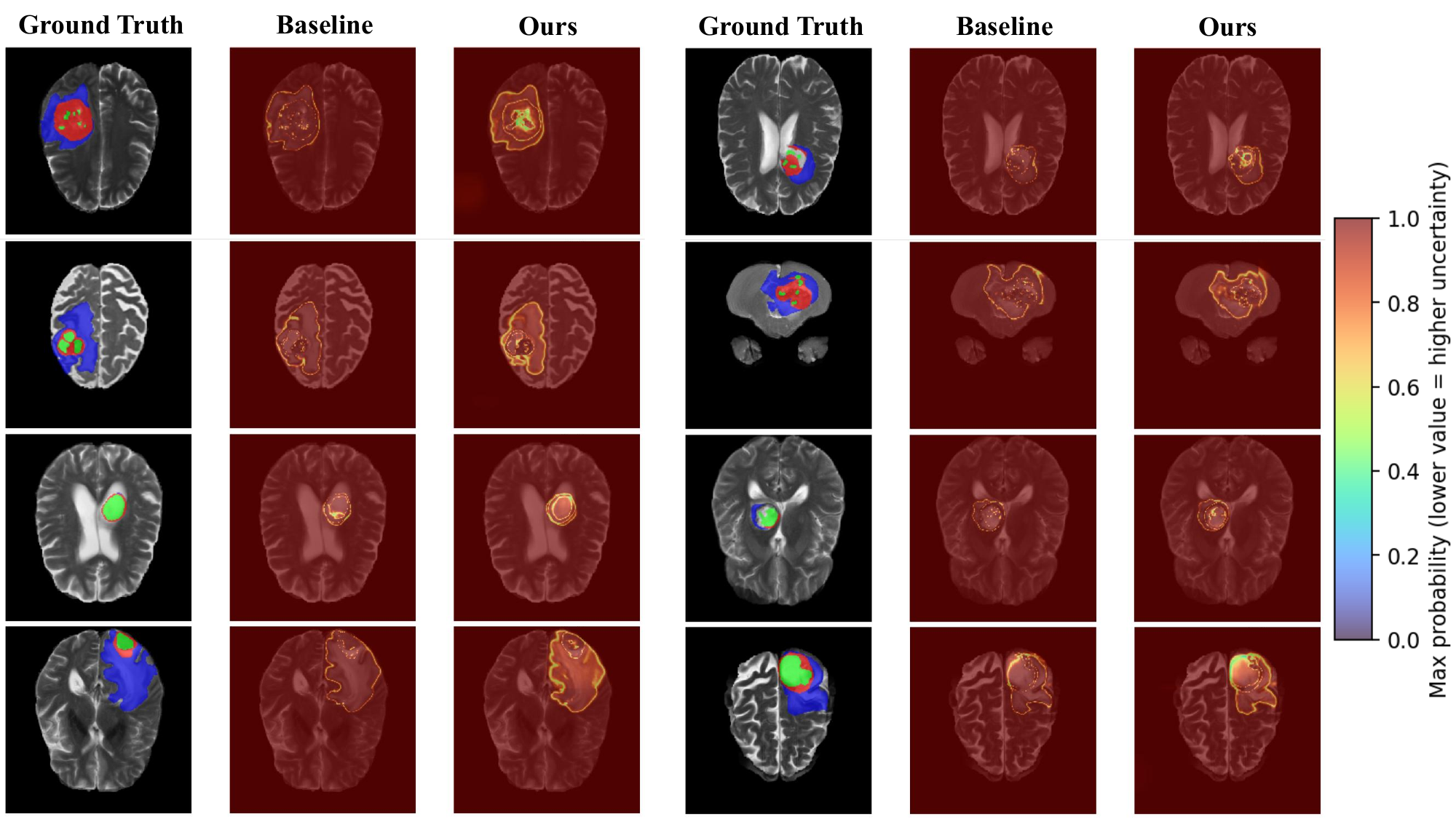}
  \caption{Visualization of softmax output probabilities. The softmax maps show that our method assigns higher uncertainty in boundary areas between classes, reflecting the inherent ambiguity and label noise present in these regions, demonstrating the effectiveness of our fuzzy uncertainty modeling in handling boundary ambiguity.}
  \label{fig:heat_map}
\end{figure*}
